\documentclass[sigconf]{acmart}

\setcopyright{rightsretained}

\acmConference[KDD'18 Deep Learning Day]{ACM KDD'18 Deep Learning Day}{August 2018}{London, UK}
\acmYear{2018}
\copyrightyear{2018}

\usepackage{booktabs} 
\usepackage{subcaption}
\usepackage{threeparttable}

\newcommand{\unit}[1]{\,\mathrm{#1}}
\usepackage[T1]{fontenc}  
\usepackage{textcomp}     

\newcommand{\vecx}{\mathbf{x}}
\newcommand{\vecy}{\mathbf{y}}
\newcommand{\vecv}{\mathbf{v}}
\newcommand{\vecw}{\mathbf{w}}
\newcommand{\vecb}{\mathbf{b}}

\newcommand{\fnast}{\textasteriskcentered}
\newcommand{\fndag}{\textdagger}
\newcommand{\fnddag}{\textdaggerdbl}
\newcommand{\fnsec}{\textsection}

\begin{document}

\title{End-to-end Deep Learning from Raw Sensor Data:\\ Atrial Fibrillation Detection using Wearables}

\author{Igor Gotlibovych}
\affiliation{%
	\institution{Jawbone Health}
	\city{London}
	\country{UK}
}
\email{igor.gotlibovych@gmail.com}

\author{Stuart Crawford}
\affiliation{%
	\institution{Jawbone Health}
	\city{San Francisco}
	\state{California}
	\country{USA}
}
\email{scrawford@jawbone.com }

\author{Dileep Goyal}
\affiliation{%
	\institution{Jawbone Health}
	\city{San Francisco}
	\state{California}
	\country{USA}
}
\email{dgoyal@jawbone.com }

\author{Jiaqi Liu}
\affiliation{%
	\institution{Jawbone Health}
	\city{San Francisco}
	\state{California}
	\country{USA}
}
\email{jliu@jawbone.com }

\author{Yaniv Kerem}
\affiliation{%
	\institution{Jawbone Health}
	\city{San Francisco}
	\state{California}
	\country{USA}
}
\email{ykerem@jawbone.com }

\author{David Benaron}
\affiliation{%
	\institution{Jawbone Health}
	\city{San Francisco}
	\state{California}
	\country{USA}
}
\email{dbenaron@jawbone.com }

\author{Defne Yilmaz}
\affiliation{%
	\institution{UCSF}
	\city{San Francisco}
	\state{California}
	\country{USA}
}
\email{defne.yilmaz@ucsf.edu}

\author{Gregory Marcus}
\affiliation{%
	\institution{UCSF}
	\city{San Francisco}
	\state{California}
	\country{USA}
}
\email{greg.marcus@ucsf.edu}

\author{Yihan (Jessie) Li}
\authornote{corresponding author}
\orcid{0000-0002-3492-8873}
\affiliation{%
	\institution{Jawbone Health}
	\city{London}
	\country{UK}
}
\email{jessieli@jawbone.com}

\renewcommand{\shortauthors}{I. Gotlibovych et al.}

\begin{abstract}
	We present a convolutional-recurrent neural network architecture with long short-term memory
	for real-time processing and classification of digital sensor data. 
	The network implicitly performs typical signal processing tasks such as filtering and peak detection, and learns time-resolved embeddings of the input signal.
	
	We use a prototype multi-sensor wearable device to collect over $180\unit{h}$ of photoplethysmography (PPG) data sampled at $20\unit{Hz}$, of which $36\unit{h}$ are during atrial
	fibrillation (AFib).
	
	We use end-to-end learning to achieve state-of-the-art results in detecting AFib from raw PPG data. 
	For classification labels output every $0.8\unit{s}$, we demonstrate an area under ROC curve of $0.9999$, with false positive and false negative rates both below $2\times 10^{-3}$.
	
	This constitutes a significant improvement on previous results utilising domain-specific feature engineering, such as heart rate extraction, and brings
	large-scale atrial fibrillation screenings within imminent reach.
	
\end{abstract}



\maketitle

\section{Introduction}
\subsection{Atrial fibrillation}\label{sec:afib}
Atrial fibrillation (AFib) is a condition characterised by an irregular and often rapid heartbeat due to abnormalities in the heart's electrical activity.
It affects between 2--3\% of the population \cite{ballAtrialFibrillationProfile2013}, yet as many as 50\% of cases remain undiagnosed for 5+ years \cite{zoni-berissoEpidemiologyAtrialFibrillation2014}.
It causes a range of complications, including stroke \cite{stewartPopulationbasedStudyLongterm2002}, yet can be managed successfully if
diagnosed early \cite{january2014AHAACC2014}.
Despite significant interest, AFib detection is complicated by several factors:
First, it typically relies on an electrocardiogram (ECG) recorded in a hospital setting. Second,
due to intermittent nature of the condition, patients may not exhibit any symptoms at the time of the recording, and may require prolonged monitoring.
Finally, diagnosis is made by trained cardiologists, and screening efforts are thus difficult to scale to the larger population.

\subsection{Wearable devices for AFib diagnostics}\label{sec:wearables}
Photoplethysmography (PPG) has been proposed as a lower-cost alternative to ECG for the purpose of AFib detection.
PPG heart rate monitors have already found wide-spread use in wearable consumer devices such as fitness trackers and smart watches.
Unlike ECG, PPG measures changes in the intensity of light reflected by the user's skin due to varying volume and oxygenation of blood in the capillaries \cite{allenPhotoplethysmographyItsApplication2007}. In a recent study \cite{tisonPassiveDetectionAtrial2018}, 
it was shown that heart rate readings from an Apple Watch could be useful in detecting AFib.

In the present study, we develop a neural-network-based algorithm to detect AFib from raw PPG signal. The sensor signal is provided by a wrist-worn prototype fitness tracker device, and sampled continuously at $20\unit{Hz}$. By training a neural network to perform all stages of feature extraction and classification, we achieve performance far superior to what is possible from heart rate features alone.

\section{Data}\label{sec:data}
The intermittent nature of AFib presents significant challenges to data collection. We collaborate with the University of California, San Francisco (UCSF) Division of Cardiology to record a range of signals as patients undergo cardioversion - a medical procedure that restores normal sinus rhythm (NSR)
in patients with AFib through electric shocks. The procedure is performed under conscious sedation, limiting both the patient's discomfort and movement.
Participants are of a diverse demographic, covering a range of ages (37--85 years), skin types (I--V on the
Fitzpatrick scale \cite{fitzpatrickValidityPracticalitySunReactive1988}),
races (77\% white), and both genders (71\% male). 
Cardiologist-reviewed ECGs are used to infer the ground truth labels before and after cardioversion.
We exclude a minority of regions labelled by experts as other arrhythmias,
and exclude one patient from the test set due to insufficient ECG data during recurring AFib episodes post-cardioversion.
In addition, we record data from volunteers with no known arrhythmias during sleep outside the hospital setting. We assume that these internal recordings do not contain episodes of atrial fibrillation.
We do not exclude recordings or parts thereof based on PPG signal quality, and allow for possibility of mislabelled regions in the training data due to insufficient ECG coverage.
Table \ref{tab:data} summarizes the data used for algorithm development and testing.

\begin{table}
	\begin{threeparttable}
		\centering
		\caption{Train and test data.}\label{tab:data}
		\renewcommand{\thefootnote}{\fnsymbol{footnote}}
		\begin{tabular}{lrrrr}
			\toprule
			& \textbf{source} & \textbf{subjects} & \textbf{rhythm} & \textbf{duration [h]}\\
			\midrule
			\textit{train} & UCSF\tnote{\fnast}   &  29 & AFib\tnote{\fnddag} & 30 \\
			&              &      & NSR\tnote{\fnsec} & 15 \\
			& internal\tnote{\fndag} & 13  & NSR\tnote{\fnsec} & 100 \\
			\hline
			\textit{test}  & UCSF\tnote{\fnast}    & 7   & AFib\tnote{\fnddag} & 6 \\
			&     	      &	     & NSR\tnote{\fnsec} & 3 \\
			& internal\tnote{\fndag} & 4  & NSR\tnote{\fnsec} & 25 \\
			\bottomrule	
		\end{tabular}
		\begin{tablenotes}
			\item [\fnast] patients undergoing cardioversion, awake
			\item [\fndag] volunteers with no known arrhythmias, asleep
			\item [\fnddag] atrial fibrillation
			\item [\fnsec] normal sinus rhythm
			
		\end{tablenotes}
	\end{threeparttable}
\end{table}

Results presented here are based on approximately $180\unit{h}$ of data, of which $36\unit{h}$ are AFib. This is equivalent to approximately $10^7$ raw samples,
or $10^6$ individual heartbeats. Using raw data maximises the information available for classification, and opens up numerous possibilities for generating augmented data, as discussed
in Section \ref{sec:training}.

\section{Classifying Raw PPG Signals}\label{sec:model}
The bottom panel in Fig. \ref{fig:inception} shows a typical PPG signal as the patient transitions from  AFib to a normal sinus rhythm.
Changes in the amplitude and periodicity of the signal are apparent, but presentation varies over time and between patients.
By using a suitable heartbeat segmentation algorithm, it is possible to extract a range of features describing variability in periods and amplitudes, as well as morphology, of individual heartbeats. Insets of Fig. \ref{fig:inception} (top panel) illustrate the value of this approach, yet choosing relevant features is a non-trivial task.
Real-world issues such as signal discontinuities and noise from a range of sources further complicate the classification problem. It is common practice to pre-process the signal, exclude noisy regions using a separate criterion, or introduce an additional label for such regions.
Importantly, the information content in noisy signals per unit time may vary, which must be reflected in the classifier output.

\subsection{Related work}
A range of timeseries classification techniques have been proposed \cite{wangTimeSeriesClassification2016}, with deep learning gaining increasing traction \cite{bagnallGreatTimeSeries2017}.
Recent work on classifying PPG signals can be broadly divided into
time-domain heart rate approaches relying on heartbeat segmentation \cite{nematiMonitoringDetectingAtrial2016}, and
frequency-domain approaches generating features through Fourier or wavelet transforms \cite{shashikumarDeepLearningApproach2017}.
Classification of ECG signals has received significant attention, with deep learning approaches employed almost exclusively in recent work \cite{rajpurkarCardiologistLevelArrhythmiaDetection2017,zihlmannConvolutionalRecurrentNeural2017,xiaDetectingAtrialFibrillation2018,shashikumarDetectionParoxysmalAtrial2018}.

Our work on classifying medical sensor signals benefits from the many advances made using convolutional and recurrent neural networks in the domains of audio labelling and synthesis \cite{hannunDeepSpeechScaling2014, sakFastAccurateRecurrent2015, oordWaveNetGenerativeModel2016}, and image recognition
\cite{lecunBackpropagationAppliedHandwritten1989,
	lecunGradientbasedLearningApplied1998,
	heDeepResidualLearning2016,
	russakovskyImageNetLargeScale2015,
	srivastavaTrainingVeryDeep2015}.

\subsection{Convolutional-recurrent architecture}\label{sec:architecture}
\begin{figure}
	\centering
	\includegraphics[width=\columnwidth]{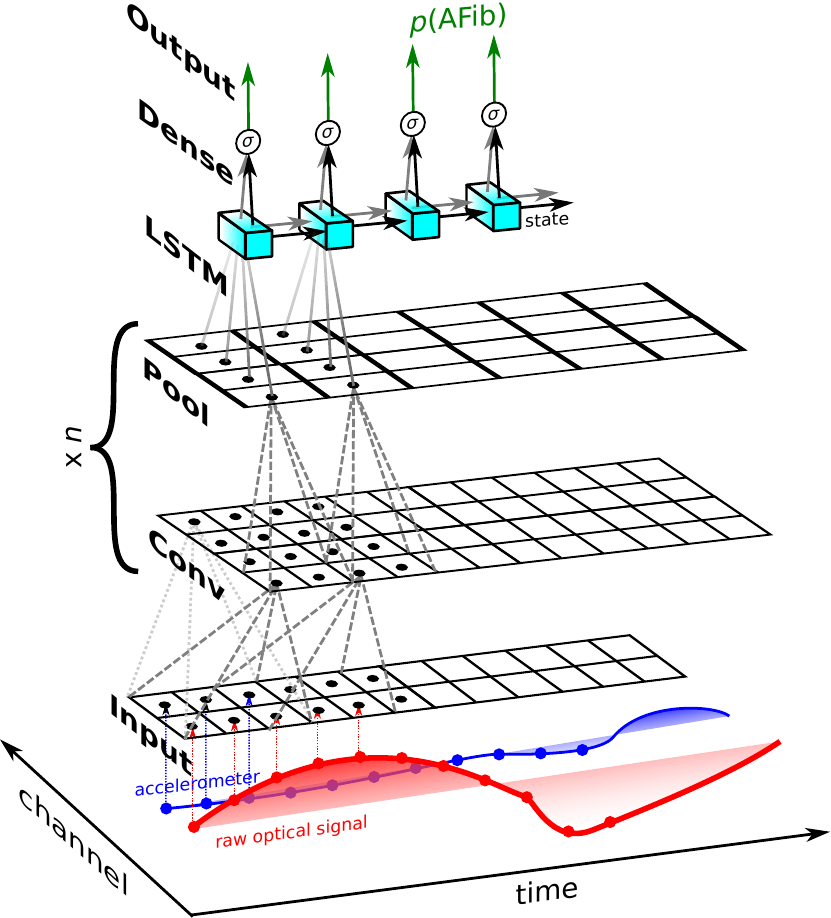}
	\caption{A convolutional-recurrent architecture for classification of raw time-series data.
		While the receptive field of each neuron in the convolutional (Conv) layers is well defined,
		the recurrent long short-term memory (LSTM) layer can learn variable-length correlations.
	}
	\label{fig:architecture}
\end{figure}

\begin{figure*}
	\centering
	\includegraphics[width=\textwidth]{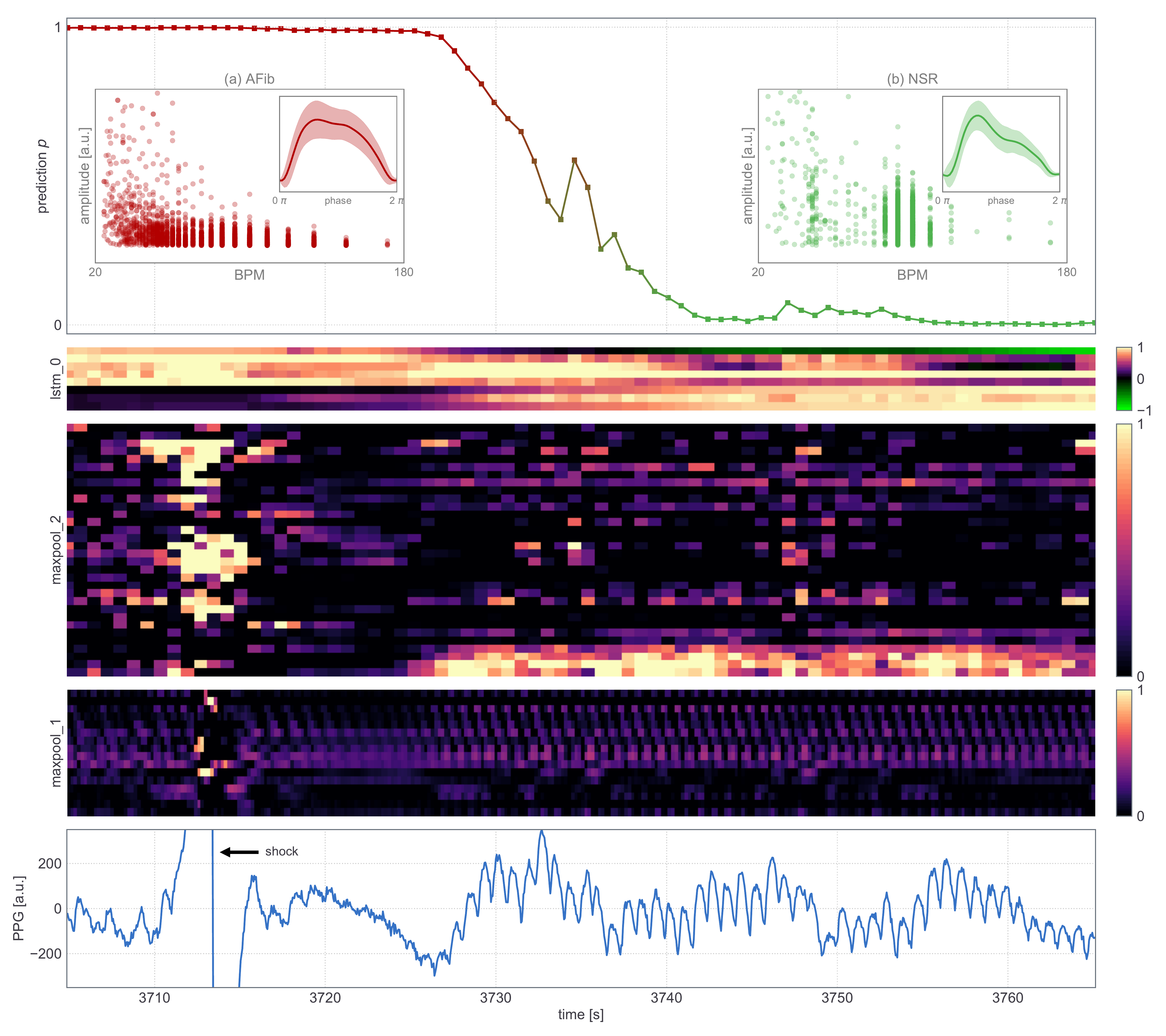}	
	\caption{
		Real-time labelling of AFib vs NSR from raw PPG signal during cardioversion. Bottom to top: 
		PPG signal; time-aligned activations in intermediate layers of the unrolled network; output probability $p$ (see text).
		Insets show typical variations in heart rates (BPM), amplitudes, and PPG morphology for individual heartbeats during (a) AFib and (b) NSR.
		Details on visualizing individual activations are given in Appendix \ref{apx:neurons}.
	}
	\label{fig:inception}
\end{figure*}

To overcome the issues outlined above, we propose an end-to-end model mapping the inputs to a sequence of calibrated, instantaneous probabilities.
The model is based on the convolutional-recurrent neural network architecture
shown schematically in Fig. \ref{fig:architecture}.

Our input is a sequence of samples $x_{t_i}$, recorded at times $t_i$. The corresponding sampling frequency is $f_x = (t_{i+1}-t{i})^{-1}$.
We seek to predict the sequence of probabilities
$$
p_{\tau_j} = P(\textrm{AFib at}\,\tau_j | x_{t_i \leq \tau_j})
$$
Notice that our approach allows for the output $p$ to depend on
all previous values of $x_{t_i}$.
Convolutional layers \cite{lecunConvolutionalNetworksApplications2010} with ReLU non-linearities \cite{jarrettWhatBestMultistage2009, glorotDeepSparseRectifier2011} extract multiple new features each layer, based on a receptive field of fixed length.
\footnote{the receptive field could be expanded significantly, e.g. using dilated convolutions as in \cite{oordWaveNetGenerativeModel2016}}
Convolution kernels can be seen as digital signal filters, and remove the need for hand-engineered signal processing operations.
Max-pooling \cite{hutchisonEvaluationPoolingOperations2010} is commonly used in deep convolutional neural networks,
and in the context of signal processing it can be interpreted as a down-sampling operation.\footnote{another way to down-sample the signal is through strided convolutions \cite{dumoulinGuideConvolutionArithmetic2016}}
A variable receptive field of each output is achieved by applying a long short-term memory (LSTM) recurrent layer \cite{hochreiterLongShorttermMemory1997, gersLearningForgetContinual2000},\footnote{in theory, LSTM state will depend on all previous $x_{t_i}$, though practical limitations exist \cite{bengioLearningLongtermDependencies1994}} followed by a single dense layer with sigmoid activation for the final output $p$.

The convolutional-recurrent architecture has further practical advantages:
the sequence lengths used for training or prediction are flexible,
and a real-time implementation is possible on a range of platforms.

The output frequency $f_p = (\tau_{j+1}-\tau_{j})^{-1}$ is constrained to the divisors of $f_x$.
The overall down-sampling ratio we use is $f_p/f_s = 1/16$,
i.e. a new label is output every $0.8\unit{s}$ for an input signal sampled at $20\unit{Hz}$.

Our implementation uses proven open-source libraries \cite{tensorflow2015-whitepaper, chollet2015keras, scikit-learn}.
The model hyperparameters are chosen through cross-validation.
\footnote{like the train-test split, all cross-validation splits are by subject to obtain an unbiased estimate of model performance}
We find that our model is robust over a wide range of hyperparameters, with overfitting largely controlled by data augmentation at training time, as described in the following section.

\subsection{Model training}\label{sec:training}
We seek to minimize the binary cross-entropy loss function, summed over all outputs.
The loss function is adjusted for class imbalance \cite{haiboheLearningImbalancedData2009}.

Our model contains ca. 10000 trainable parameters, and we follow best practices to improve convergence, reduce training time, and control over-fitting.
These include weights initialization \cite{glorotUnderstandingDifficultyTraining2010, sutskeverImportanceInitializationMomentum2013}, batch normalization between layers \cite{ioffeBatchNormalizationAccelerating2015}, dropout in the LSTM layer \cite{galTheoreticallyGroundedApplication2016}, and the choice of optimizer \cite{kingmaAdamMethodStochastic2014}.

We train our network on mini-batches of fixed-length subsequences of the training data.
The LSTM state is initialized at random for each example, and example length is chosen to allow the learning of long-range dependencies.
Each epoch, we perform random augmentation of the training batches.
Data augmentation has become a standard technique for training neural networks for image classification \cite{chatfieldReturnDevilDetails2014}, audio tasks \cite{cuiDataAugmentationDeep2014}, and other timeseries applications \cite{guennecDataAugmentationTime2016, umDataAugmentationWearable2017}.
Using raw data allows us to identify domain-specific heuristics for data augmentation, and thus account for e.g. variations in user skin tone and varying light conditions. We randomly offset selected examples within the raw training signals (random cropping), and apply scaling, additive shifts and random Gaussian noise with random amplitudes per example.
Random augmentation proves crucial to obtaining a model with superior performance on real-world signals. 

\begin{figure}
	\centering
	\includegraphics[width=0.85\columnwidth]{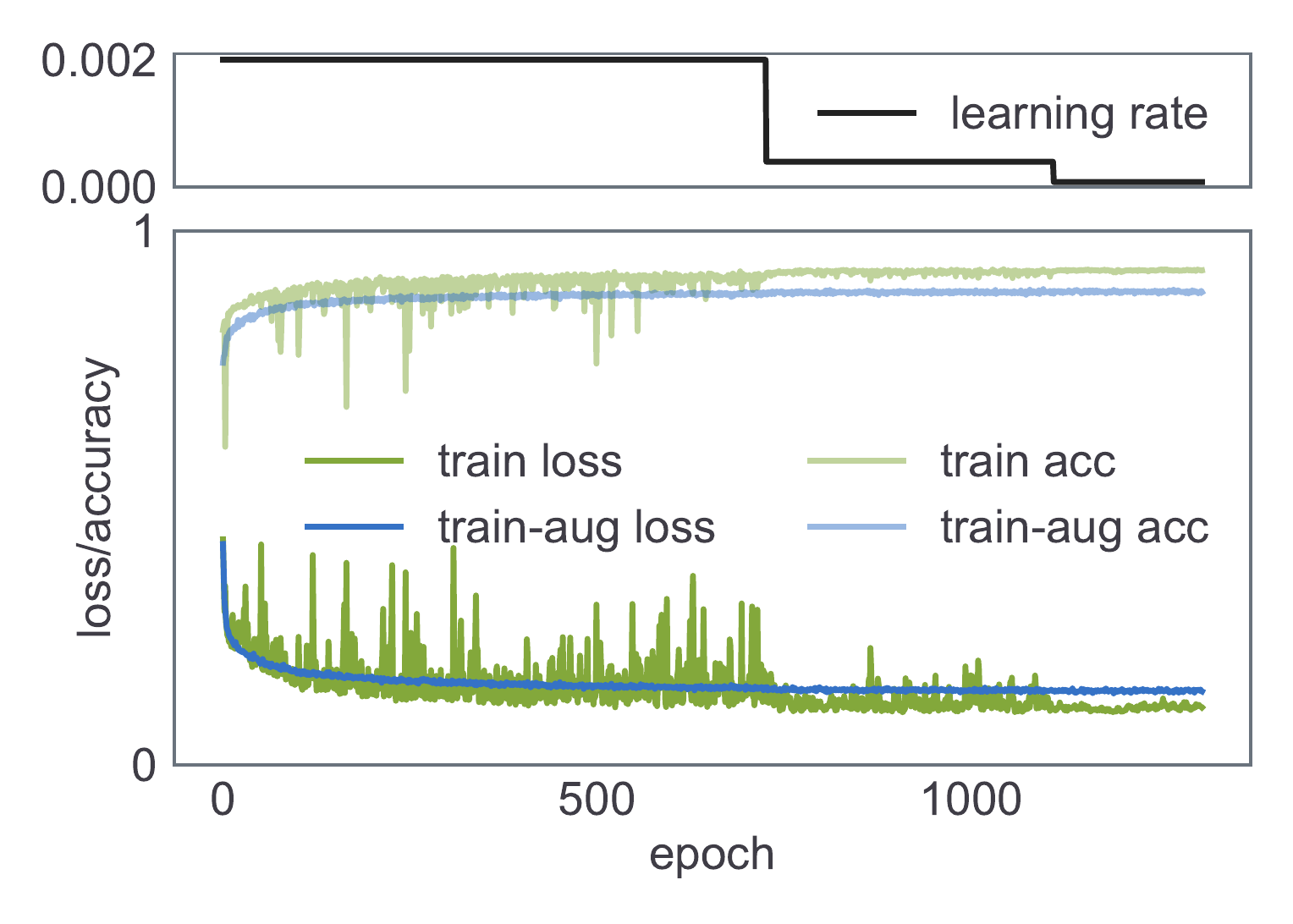}	
	\caption{Learning curves (bottom) and learning rate annealing (top) with random augmentation}
	\label{fig:learning}
\end{figure}

To monitor convergence, we use a validation set of non-overlapping, unaugmented subsequences,
reducing the learning rate every time the validation loss stops decreasing,
as seen in Figure \ref{fig:learning}.

We generally achieve better performance on unaltered validation data compared to randomly augmented training batches. Similarly, we find that the performance of the trained model on the test set is unaffected by the presence of noisy recordings in the training set, and is robust to the
presence of some mislabelled training examples. This is especially important given
the limitations of our dataset explained in Section \ref{sec:data}.

\section{Results}\label{sec:results}
\subsection{Classifier performance}\label{sec:roc}
\begin{figure}
	\centering
	\includegraphics[width=0.85\columnwidth]{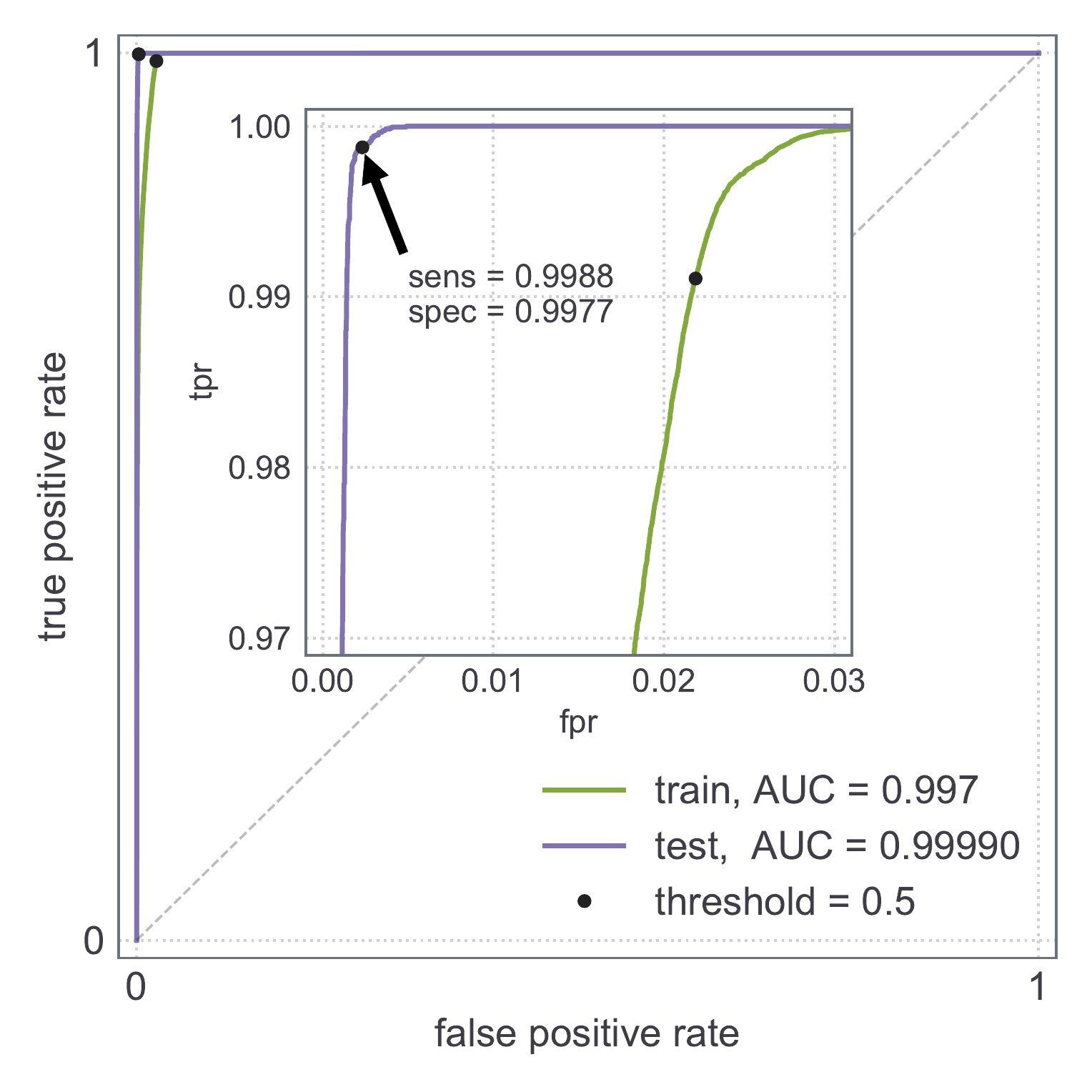} \\
	\caption{ROC of the trained model. Highlighted point corresponds to a probability threshold of $0.5$.}
	\label{fig:roc}
\end{figure}
We evaluate performance of our model on the test set of recordings, as summarised in Table \ref{tab:data}. We use raw sequence labels at $1.25\unit{Hz}$.
Figure \ref{fig:roc} shows the receiver operating characteristic (ROC) curve for our probability predictions, for both train and test data.
On the test set, we achieve AFib vs NSR classification with a specificity and sensitivity of 0.998 and 0.999, respectively, at a probability threshold of 0.5. This corresponds to a false positive rate of $2\times 10^{-3}$ and a false negative rate of $1\times10^{-3}$.
The probability output is well calibrated, with a Brier score \cite{brierVerificationForecastsExpressed1950}  of 0.002.
As noted above, we have chosen to not exclude recordings with low signal quality, nor have we excluded a recording with suspected heart rhythm changes inbetween ECG spot checks from the training data - the ability to train a highly accurate classifier despite the likely presence of mis-labelled data is important given the nature of physiological signals.

In large-scale screening applications, we expect a low false positive rate to be of key importance:
not only is the fraction of individuals with AFib small, they are expected to exhibit AFib for a fraction of the time, with episodes varying in duration and frequency.
\footnote{the total fraction of time spent in AFib by a given individual is known as the \emph{AFib burden}; we are not aware of a study describing the distribution of burdens nor episode lengths}
Considering the recordings from (presumed) healthy individuals during sleep only, we observe a
false positive rate of $0.0016$ at the same probability threshold.

\subsection{Learned signal filtering}\label{sec:dsp}

While the meaning of individual network weights is difficult to interpret,
we can identify one specific task our network learns through training: that of signal filtering.
The first convolutional layer can be seen as a bank of finite impulse response (FIR) filters, and
we find that they adapt to perform high-pass filtering, with DC attenuations ranging from $-37\unit{dB}$ to $-64\unit{dB}$.
Thus, our approach removes the need for  signal pre-processing, and the attenuation is consistent with the range of DC amplitudes seen in training.

\subsection{Neuron function}\label{sec:neurons}
Visualisation and interpretation of the function of individual neurons in convolutional \cite{olahBuildingBlocksInterpretability2018}
and recurrent \cite{carterExperimentsHandwritingNeural2016} neural networks is an area of active research.
Figure \ref{fig:inception} shows time-resolved activations after two intermediate max-pooling layers, as well as the LSTM hidden state, time-aligned with the input signal. 
We can see how a number of neurons appear to specialize in tasks such as detecting peaks in layer \texttt{maxpool\_1}, tracking persistent heart rhythm in layer \texttt{maxpool\_2}, and finally encoding presence of AFib and/or NSR in the LSTM layer.
It is interesting to note the time offset between transitions in individual LSTM hidden state values,
and also the robust behaviour in the presence of input signal discontinuities and variable signal-to-noise ratios.

\subsection{Heart rhythm embeddings}\label{sec:clustering}
\begin{figure}
	\centering
	\includegraphics[width=\columnwidth]{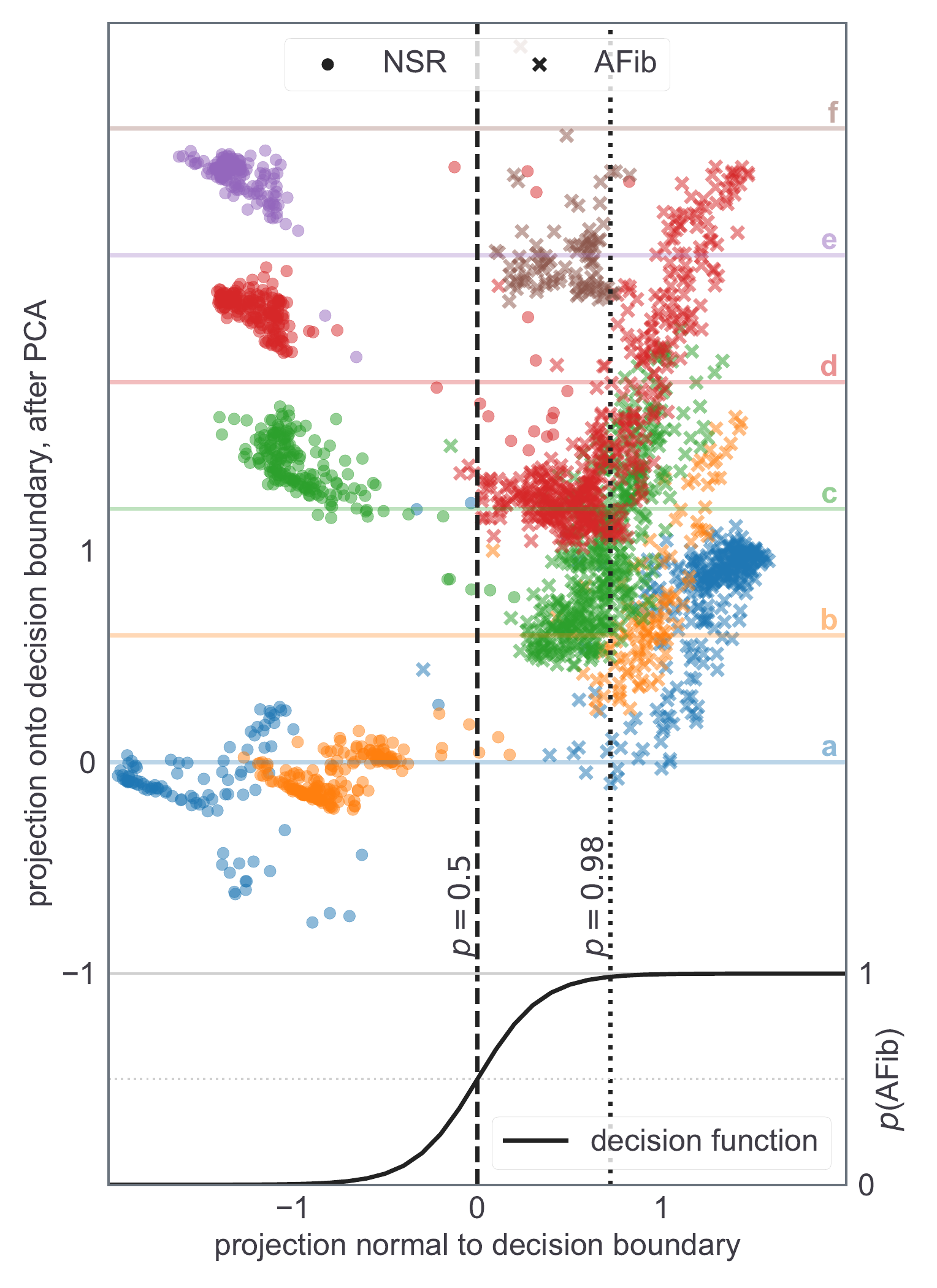}
	\caption{
		Heart rhythm embeddings, shown as 2D projections of LSTM hidden state vectors.
		Each point corresponds to one temporal step in the output sequence.
		For visual clarity, one in every 15 time steps is shown.
		Marker shape indicates the ground truth label; colours and letters correspond to unique patients.
		Projections were obtained as described in Appendix \ref{apx:2D}.
	}
	\label{fig:cluster}
	\label{fig:cluster_post}
\end{figure}
The hidden state of the LSTM layer
can be interpreted as a time-dependent latent-space embedding of the underlying heart rhythm. We visualize 2D projections of these vectors
for a range of patients in Figure \ref{fig:cluster}.
While our network learns a global decision boundary (shown for $p=0.5$),
we can see that the heart rhythm embeddings for both AFib and NSR vary between patients. 
We propose that standard unsupervised clustering techniques \cite{murphyMachineLearningProbabilistic2012},
applied to heart rhythm embeddings produced by our network, can further improve classification accuracy.
More importantly, we envision being able to detect heart rhythm anomalies in individual subjects as outliers in the latent space, and extending our approach to other heart rhythm anomalies in the future.

\section{Conclusions}\label{sec:conclusions}
In this article, we have demonstrated how applying best practices from domains such as image classification and natural language processing to the hitherto under-explored application area of real-time sensor data classification yields state-of-the-art results in PPG-based diagnostics of atrial fibrillation.
We show that digital signal pre-processing can be learned by a suitably chosen neural network architecture,
in a way that easily generalises to a multi-sensor, multi-channel setting.
By interpreting intermediate outputs of a pre-trained neural network as latent-space embeddings of the physiological signal, we can further personalize diagnostics through unsupervised learning.

One aspect that could affect real-world performance of the model is the minimum duration of an isolated AFib episode that we are able to detect. Our experiments with synthetic data
\footnote{obtained by splicing regions with different heart rhythms}
show that minimum to be between 20--200 s, with a strong variation between patients, and dependent on signal-to-noise ratios. We believe that using synthetic data at training time may improve this further - concurrently, our ongoing data collection and labelling efforts focus on capturing a variety of real-world episodes.

While we have made every effort to train a robust and generalizable model, we have only accessed performance on data collected either in a hospital setting or during sleep.
It remains to be seen how other factors such as motion and differing demographics affect the results.
At the same time, we are confident that our approach will be applicable to new and larger datasets.

Three main issues have thus far precluded large-scale preventive diagnostics of AFib: the cost and availability of ECG monitoring devices, the episodic nature of the condition, and the need for expert review. By combining low-cost wearable sensors with deep learning algorithms, we pave the way to real time detection of atrial fibrillation in millions of users.

\begin{acks}
	We are grateful to Vasilis Kontis and David Grimes for their constructive comments on the
	manuscript.
\end{acks}

\appendix

\section{Visualising Neurons}\label{apx:neurons}
In Figure \ref{fig:inception}, we show activations of intermediate-layer neurons over time. We aim to show groups of neurons that learn similar functions.
ReLU, and therefore max-pooling activations, are in the range $\left[0, \infty \right)$, while the hidden state values of an LSTM are in the range $\left[-1, 1\right]$.
To better visualise the function of our network, we order individual channels $i$ in each layer $l$ by similarity of the activation 
timeseries $a_{it}^{(l)}$,  where $t$ denotes the time index. We use the optimal leaf ordering \cite{bar-josephFastOptimalLeaf2001} obtained through hierarchical agglomerative clustering \cite{mullnerModernHierarchicalAgglomerative2011} with a suitable pairwise distance function. We find that the distance metric $d^{(l)}_{ij} = 1-\left|\mathrm{corr}\left( a_{it}^{(l)}, a_{jt}^{(l)}\right)\right|$, computed over all times, yields good results.
For the LSTM state, we invert the sign for channels with predominantly negative values (this is equivalent to flipping the sign of some weights to yield an equivalent network).

\section{Visualising vector embeddings}\label{apx:2D}
In Figure \ref{fig:cluster}, we visualise multi-dimensional vector embeddings by projecting them onto 2D. This is done in a way that preserves
the decision boundary, as defined by $\vecx \cdot \vecw + \vecb = 0$ for embeddings $\vecx \in \mathbb{R}^n$, and parameters of the simple linear classifier
$\vecw, \vecb \in \mathbb{R}^n$.
The corresponding logistic regression decision function is given by $p(\vecx) = \sigma(\vecx \cdot \vecw + \vecb)$,
with sigmoid activation $\sigma(a) = \left(1+e^{-a}\right)^{-1}$.
$\vecw$ and $\vecb$ are learned by the output layer of the network during training.

To obtain 2D projections $\vecy = (y_0, y_1)$, we write $y_0 = \vecx\cdot \hat \vecw + \hat \vecb$ and $y_1 = \mathrm{PCA}_0 \left( \vecx - \hat \vecw y_0 \right)$.
We use the notation $\hat \vecw = \vecw/{|\vecw|}$, $\hat \vecb = \vecb/{|\vecw|}$ for normalized vectors, and $\textrm{PCA}_n(\vecv)$ denotes the $n$th principal component of $\vecv$.

\bibliographystyle{ACM-Reference-Format}

\end{document}